\documentclass[letterpaper]{article} 
\usepackage{aaai2026}  
\usepackage{times}  
\usepackage{helvet}  
\usepackage{courier}  
\usepackage[hyphens]{url}  
\usepackage{graphicx} 
\usepackage{amsmath} 
\usepackage{natbib}  
\usepackage{caption} 
\usepackage{algorithm}
\usepackage{algorithmic}
\usepackage{amssymb}
\usepackage{booktabs}
\usepackage{pgffor} 
\usepackage{caption}
\usepackage{xcolor}
\usepackage{multirow}
\usepackage{color}
\usepackage{graphicx}
\usepackage{xcolor}
\usepackage{subfigure}

\definecolor{ForestGreen}{RGB}{34,139,34}

\usepackage{newfloat}
\usepackage{listings}
\DeclareCaptionStyle{ruled}{labelfont=normalfont,labelsep=colon,strut=off} 
\floatstyle{ruled}
\newfloat{listing}{tb}{lst}{}
\floatname{listing}{Listing}
\newcommand{\name}{\texttt{\textbf{Robust-R1}}}

\title{\name: Degradation-Aware Reasoning for Robust Visual Understanding}

\author{
    Jiaqi Tang\textsuperscript{\rm 1}\equalcontrib,
    Jianmin Chen\textsuperscript{\rm 2}\equalcontrib, 
    Wei Wei\textsuperscript{\rm 2}\footnotemark[2],
    Xiaogang Xu\textsuperscript{\rm 3},
    Runtao Liu\textsuperscript{\rm 1},\\
    Xiangyu Wu\textsuperscript{\rm 4},
    Qipeng Xie\textsuperscript{\rm 1},
    Jiafei Wu\textsuperscript{\rm 5},
    Lei Zhang\textsuperscript{\rm 2},
    Qifeng Chen\textsuperscript{\rm 1}\thanks{Corresponding Author: Qifeng Chen; Co-corresponding Author: Wei Wei.}
}
\affiliations{
    \textsuperscript{\rm 1}Hong Kong University of Science and Technology\\
    \textsuperscript{\rm 2}Northwestern Polytechnical University\\
    \textsuperscript{\rm 3}Chinese University of Hong Kong\\
    \textsuperscript{\rm 4}Nanjing University of Science and Technology\\
    \textsuperscript{\rm 5}University of Hong Kong \\
    \url{cqf@ust.hk}, weiweinwpu@nwpu.edu.cn
}

\usepackage{bibentry}

\begin{document}

\maketitle

\begin{abstract}
Multimodal Large Language Models struggle to maintain reliable performance under extreme real-world visual degradations, which impede their practical robustness.
Existing robust MLLMs predominantly rely on implicit training/adaptation that focuses solely on visual encoder generalization, suffering from limited interpretability and isolated optimization.
To overcome these limitations, we propose \name, a novel framework that explicitly models visual degradations through structured reasoning chains.
Our approach integrates: (i) supervised fine-tuning for degradation-aware reasoning foundations, (ii) reward-driven alignment for accurately perceiving degradation parameters, and (iii) dynamic reasoning depth scaling adapted to degradation intensity.
To facilitate this approach, we introduce a specialized 11K dataset featuring realistic degradations synthesized across four critical real-world visual processing stages, each annotated with structured chains connecting degradation parameters, perceptual influence, pristine semantic reasoning chain, and conclusion.
Comprehensive evaluations demonstrate state-of-the-art robustness: \name\ outperforms all general and robust baselines on the real-world degradation benchmark R-Bench, while maintaining superior anti-degradation performance under multi-intensity adversarial degradations on MMMB, MMStar, and RealWorldQA. 
\end{abstract}

%
\begin{links}
    \link{Code}{github.com/jqtangust/Robust-R1}
    \link{Data}{huggingface.co/datasets/Jiaqi-hkust/Robust-R1}
    \link{Model}{huggingface.co/Jiaqi-hkust/Robust-R1}
    \link{Space}{huggingface.co/spaces/Jiaqi-hkust/Robust-R1}
\end{links}

\section{Introduction}
\begin{figure}[t]
    \centering
    \includegraphics[width=1\linewidth]{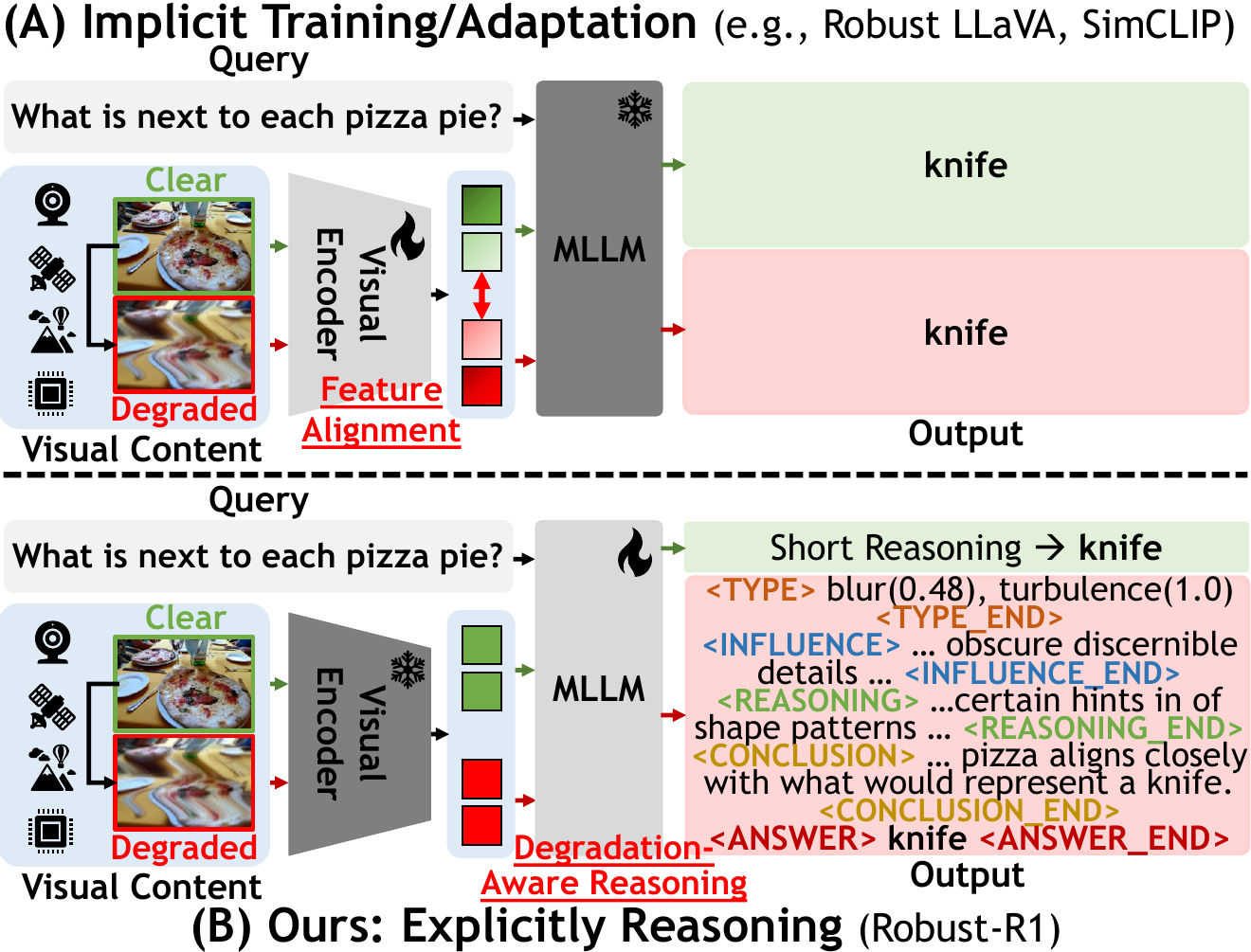}
    \caption{Comparison with other existing robustness enhancement approaches. (A) is based on implicit training/adaptation, which only considers the visual encoder feature alignment. (B) is ours, and we explicitly integrate the degradation-aware reasoning chain into MLLM.}
    \label{motivation1}
\end{figure}

Multimodal Large Language Models (MLLMs) have demonstrated remarkable capabilities in visual understanding tasks~\cite{liu2023improvedllava,NEURIPS2024_fca83589,tang2025lpoaccurateguiagent,lu2024gpt}. However, their performance degrades significantly under real-world visual degradations (e.g., noise, blur, occlusion)~\cite{malik2025robust,schlarmann2024robustclip,tang2023high,tang2024learning}, compromising reliability in practical applications. Therefore, enhancing robustness against such degradations remains a critical challenge for deploying MLLMs in uncontrolled environments~\cite{long2025robust}.

Existing approaches primarily rely on \textit{implicit training/adaptation strategies} to integrate robustness, such as adversarial training~\cite{wang2024tecoa}, robust vision-language alignment~\cite{hossain2024sim,schlarmann2024robustclip,yuan2024helpd}, or large-scale adversarial pre-training~\cite{malik2025robust}. These methods focus on fortifying visual encoders against distortions through data-centric optimization.
While effective, they suffer from two fundamental limitations (as indicated in Figure~\ref{motivation1}-A):  
\textbf{(i) Limited Interpretability:} They lack explicit mechanisms to diagnose degradation impacts on original semantic information.  
\textbf{(ii) Isolated Optimization:} They neglect the degradation-propagation relation between the visual encoder and large language model.

To overcome these limitations, we propose \name, a novel framework that explicitly models visual degradations through structured reasoning. Unlike implicit paradigms, \name\ firstly perceives degradation parameters (type and intensity), then analyzes their semantic impact on visual content, and finally reconstructs distortion-free interpretations to derive robust results. This explicit approach significantly enhances robustness while providing interpretable reasoning traces (as shown in Figure~\ref{motivation1}-B).

Our implementation comprises three core stages: \textbf{First}, we perform Supervised Fine-Tuning (SFT) to equip pretrained MLLMs with foundational degradation-aware reasoning abilities. \textbf{Second}, we design a reward function that aligns model outputs with accurate degradation parameters. \textbf{Finally}, we introduce a complementary reward function to dynamically scale the reasoning chain length according to degradation severity, ensuring optimal efficiency.

To support this approach, we construct an 11K dataset from A-OKVQA~\cite{schwenk2022aokvqabenchmarkvisualquestion}, comprising 10K training and 1K validation samples.
For each sample, we synthesize realistic degradations by simulating four key stages: acquisition $\rightarrow$ transmission $\rightarrow$ environment $\rightarrow$ postprocessing with random intensities.
We then generate structured reasoning chains that link:
(i) degradation parameters ($\mathbf{D}_d$),
(ii) their influence ($\Delta_d$),
(iii) the pristine semantic reasoning chain ($\text{T}_{\mathbf{X}}$),
and (iv) the final conclusion ($\mathbf{Y}_d$).
The complexity of these reasoning chains is dynamically scaled with the degradation intensity to balance robustness with computational efficiency.

Comprehensive evaluations demonstrate \name's superior robustness. On the real-world degradation benchmark R-Bench~\cite{li2024rbench}, \name\ achieves state-of-the-art (SOTA) performance across all degradation intensities (low, medium, and high), outperforming existing general MLLMs and robust MLLMs.
Furthermore, when subjected to adversarial degradation on general visual understanding benchmarks (MMMB~\cite{sun2025parrotmultilingualvisualinstruction}, MMStar~\cite{chen2024we}, and RealWorldQA~\cite{xai2024grok15v}), \name\ maintains significantly robust performance. It exhibits a markedly smaller performance drop compared to all baselines under multi-level degradation intensities ($25\%$, $50\%$, and $100\%$).
Our contributions are summarized as:
\begin{itemize}
    \item We propose \name, a novel approach that explicitly mitigates visual degradations in MLLMs through structured reasoning chains, providing interpretable degradation diagnostics alongside enhanced robustness.
    \item We construct a dataset of 11K samples featuring realistic degradations synthesized across four critical stages, each annotated with structured reasoning chains for degradation-aware reasoning.
    \item \name\ achieves SOTA performance on the real-world robust visual understanding benchmark (R-Bench) and demonstrates superior robustness under adversarial degradation on established general benchmarks (MMMB, MMStar, RealWorldQA), significantly outperforming existing general and robust MLLM baselines.
\end{itemize}

\section{Related Work}

\paragraph{Robust Visual Understanding}
Environmental perturbations~\cite{ma2025surgeon,fu2025co} pose persistent challenges to multimodal large language models (MLLMs), often significantly degrading their perceptual and reasoning capabilities~\cite{lu2024gpt}.
As a result, enhancing model robustness has become a critical focus in visual understanding research.
Early efforts primarily focused on adversarial training through visual encoder fine-tuning. Approaches like TeCoA~\cite{wang2024tecoa}, Sim-CLIP~\cite{hossain2024sim}, and Robust CLIP~\cite{schlarmann2024robustclip} optimized model resilience against localized distortions but faced inherent limitations: reliance on limited adversarial datasets often compromised generalization performance.
More recent approaches, such as Robust LLaVA~\cite{malik2025robust}, have sought to mitigate these issues through large-scale adversarial pre-training. Despite some success, these strategies incur substantial computational and annotation costs, limiting their scalability.

In contrast to these implicit adaptation paradigms, \name\ introduces a novel degradation-aware reasoning mechanism that explicitly enhances interpretability while improving robustness.

\paragraph{Multimodal Reasoning}
Multimodal reasoning empowers MLLMs to solve complex tasks by integrating perception, contextual understanding, and logical inference~\cite{wei2022chain}.
Prior work has made considerable progress in domains such as mathematical visual reasoning, where models are required to interpret and reason over problems involving both symbolic notations and visual elements~\cite{wang2024measuring,lu2023mathvista}. 
Subsequent research has expanded into broader visual reasoning scenarios, exemplified by frameworks like Visual CoT~\cite{shao2024visual} and V*~\cite{wu2024v}, which focus on parsing scene elements and their relational structure.

\name\ builds upon and extends this line of work by harnessing the MLLM's intrinsic reasoning capacity, pioneering its application to explicitly reason about and overcome visual distortions, thereby establishing a new paradigm for robust multimodal understanding.

\section{Methodology}
\begin{figure*}[t]
    \centering
    \includegraphics[width=1\linewidth]{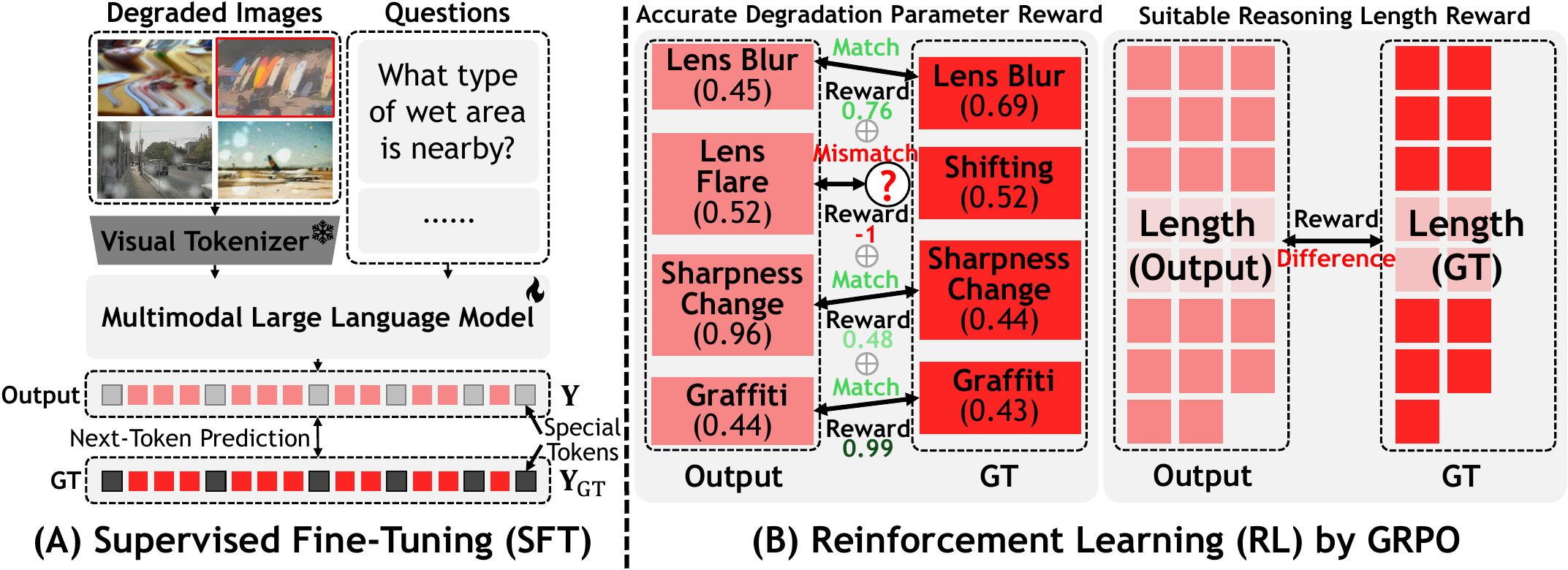}
    \caption{Overview of \name. (A) Supervised Fine-Tuning (SFT): we train the model using reasoning data to equip it with basic degradation-aware reasoning capability; (B) Reinforcement Learning (RL): we propose two reward functions to (i) align precise degradation-aware space while (ii) adaptively scaling to suitable reasoning lengths based on degradation intensity.}
    \label{method}
\end{figure*}

\paragraph{Problem Definition}
Multimodal Large Language Models (MLLMs) frequently exhibit performance degradation when processing visually corrupted inputs in real-world scenarios~\cite{xu2024low,xu2025learnable}, which undermines their interpretation accuracy.
This challenge can be represented as Eq.~\eqref{e0},
\begin{equation} \small
    \mathbf{Y}_d = \mathcal{M}_{\text{MLLM}}(\mathbf{X}_d \oplus \mathbf{P}),
    \label{e0}
\end{equation}
where $\mathbf{X}_d$ is the degraded visual input, derived as $\mathbf{X}_d = \mathcal{D}(\mathbf{X})$, with $\mathbf{X}$ as the original input and $\mathcal{D}(\cdot)$ representing the degradation function.
$\mathbf{P}$ denotes the text prompt.
$\mathcal{M}_{\text{MLLM}}(\cdot)$ denotes the original MLLM framework.
$\mathbf{Y}_d$ is the generated output under current conditions.
$\oplus$ indicates the multimodal combination operator.
To tackle this issue, we aim to develop a robust MLLM framework that satisfies:
\begin{equation}  \small
    \mathcal{M}_{\text{MLLM}}^{\text{(Robust)}}(\mathbf{X}_d \oplus \mathbf{P}) \xrightarrow{\text{approx}} \mathcal{M}_{\text{MLLM}}(\mathbf{X} \oplus \mathbf{P}),
    \label{eq:robust_objective}
\end{equation}
where $\mathcal{M}_{\text{MLLM}}^{\text{(Robust)}}(\cdot)$ denotes our enhanced model, and the approximation operator $\xrightarrow{\text{approx}}$ signifies the objective of approximating the output under pristine visual conditions.

\paragraph{Overview of Degradation-Aware Reasoning}
To address the above problem, \name~incorporates an explicit degradation-aware reasoning process that perceives degradation parameters (type and intensity), analyzes their impact on visual content, and reconstructs high-fidelity interpretations. This process is formulated as:
\begin{equation} \label{chain}  \small
\begin{aligned}
    \mathcal{M}_{\text{MLLM}}^{\text{(Robust)}}(\mathbf{X}_d \oplus \mathbf{P}) \Leftrightarrow \\ \{ 
    \mathcal{M}_{\text{p}}\bigl(\text{D}_d, \Delta_{d} \mid \mathbf{X}_d\bigr) 
 \rightarrow 
\mathcal{M}_{\text{r}}\bigl(\text{T}_{\mathbf{X}} \mid \text{D}_d, \Delta_{\text{d}}, \mathbf{X}_d, \mathbf{P}\bigr)
\\ \rightarrow
\mathcal{M}_{\text{MLLM}}\bigl( \text{Y}_d \mid (\text{T}_{\mathbf{X}}, \text{D}_d, \Delta_{d}) \oplus \mathbf{X}_d \oplus \mathbf{P}\bigr) \},
\end{aligned}
\end{equation}
where $\mathcal{M}_{\text{p}}(\cdot)$ is degradation parameters perception process, to perceive $\text{D}_d = \{\tau_d^{(i)}, s_d^{(i)}\}_{i=1}^{I}$ (types $\tau_d$ and intensities $s_d$) and their impact $\Delta_{d}$;
$\mathcal{M}_{\text{r}}(\cdot)$ reconstructs the pristine semantic representation $\text{T}_{\mathbf{X}}$ of original $\mathbf{X}$;
and original $\mathcal{M}_{\text{MLLM}}(\cdot)$ can generate the robust output $\mathbf{Y}_d$ conditioned on degradation-aware reasoning chain.

\paragraph{Workflow}
\textbf{Firstly}, to integrate degradation-aware reasoning capabilities, We first fine-tune the pretrained vision-language model to establish foundational degradation-aware reasoning capabilities (Section~\ref{subsec:1}).
\textbf{Subsequently}, We employ reinforcement learning with a dedicated reward function to align the model's perception with accurate degradation parameters ($\text{D}_d$) (Section~\ref{subsec:2}).
\textbf{Finally}, we dynamically adjust the reasoning chain length based on degradation intensity to optimize the trade-off between robustness and efficiency (Section~\ref{subsec:3}).

\subsection{Acquiring Basic Reasoning Ability} \label{subsec:1}
\paragraph{Tokenization of Reasoning Chain}
To enable structured degradation-aware reasoning, we formalize the reasoning chain using special tokens (enclosed in``$<$" and ``$>$") that segment distinct reasoning phases:
\begin{equation} \small
\begin{aligned}
\texttt{<TYPE>}&\text{D}_d\texttt{<TYPE\_END>}, \\
\texttt{<INFLUENCE>}&\Delta_d\texttt{<INFLUENCE\_END>}, \\
\texttt{<REASONING>}&\text{T}_{\mathbf{X}}\texttt{<REASONING\_END>}, \\
\texttt{<CONCLUSION>}&\mathbf{Y}_d\texttt{<CONCLUSION\_END>}, \\
\texttt{<ANSWER>}&\mathbf{Y}_d^{\text{(answer)}}\texttt{<ANSWER\_END>} \ \text{(Optional)},
\end{aligned}
\label{format}
\end{equation}
where $\mathbf{Y}_d^{\text{(answer)}}$ denotes the task-specific answer output during benchmark evaluation. This tokenization enforces a sequential reasoning flow to maintain structured output.

\paragraph{Supervised Fine-Tuning (SFT)}
We optimize model parameters $\theta$ through next-token prediction (as shown in Figure~\ref{method}-A) on the structured reasoning chain:
\begin{equation} \small
\mathcal{L}_{\text{SFT}} = -\mathbb{E}_{(\mathbf{X}_d, \mathbf{P}, \mathbf{Y}) \sim \mathcal{P}_{T}} \sum_{n=1}^{N} \log \mathcal{P}_{\theta}\left(w_n \mid w_{<n}, \mathbf{X}_d, \mathbf{P}\right),
\end{equation}
where $\text{C} = (w_n, \dots, w_N) \sim \{\mathrm{D}_d,\Delta_d \rightarrow \mathrm{T}_{\mathbf{X}} \rightarrow \mathbf{Y}_d\}$ represents the output reasoning chain. $N$ denotes the sequence length, $\mathcal{P}_{\theta}$ is the model's conditional probability distribution, $\mathcal{P}_{T}$ denotes the distribution of training data.
This optimization enables the model to acquire foundational degradation-aware reasoning ability by sequentially generating the structured reasoning chain.

\subsection{Aligning Accurate Degradation Parameters}
\label{subsec:2}
Although SFT equips the MLLM with foundational degradation-aware reasoning ability, it still lacks an accurate perception of degradation parameters (types and intensities).
As quantitatively demonstrated in Figure~\ref{ablationstat}-A (w/o $\text{D}_d$), lacking precise alignment exhibits significant deviation from practical degradation parameters, leading to limited degradation perception ability.

\paragraph{Reward for Accurate Degradation Parameters}
To achieve high-fidelity alignment, we design a reward function that directly operates in the degradation parameter space (as shown in Figure~\ref{method}-B (left)).
The reward function $r_{\text{deg}}(\mathbf{Y}, \mathbf{Y}_{\text{GT}})$ explicitly evaluates degradation parameter deviation:
\begin{equation} \small
\begin{aligned}
r_{\text{deg}}(\mathbf{Y}, \mathbf{Y}_{\text{GT}}) = \sum_{i=1}^{I} \sum_{j=1}^{J} \mathcal{\delta}(\tau_d^{(i)} = \tau_{\text{GT}}^{(j)}) \cdot \left( 1 - \left|s_d^{(i)} - s_{\text{GT}}^{(j)}\right| \right) \\ -\ \delta(\tau_d^{(i)} \neq \tau_{\text{GT}}^{(j)}),
\end{aligned}
\end{equation}
where $\delta(\cdot)$ denotes the Kronecker delta function~\cite{web_reference}.
This formulation specifically: (1) penalizes type mismatches with $-1$ reward; (2) rewards type matches proportionally to intensity accuracy ($1 - |\Delta s|$); and (3) aggregates rewards across all instances ($i=1,\dots, I$ and $j=1,\dots, J$).

\begin{figure}
    \centering
    \includegraphics[width=1\linewidth]{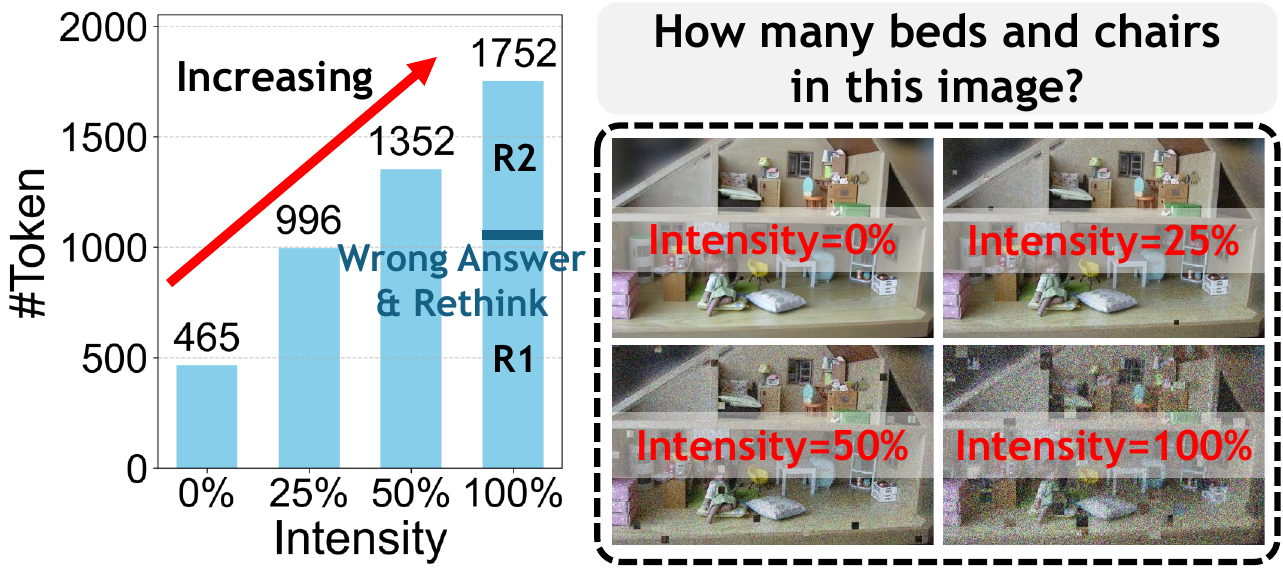}
    \caption{Correlation between degradation intensity and reasoning chain length on Seed-1.5-VL~\cite{guo2025seed1}. Higher degradation intensities require longer chains to maintain accuracy, even multi-step reasoning.}
    \label{length}
\end{figure}
\subsection{Scaling to Suitable Reasoning Length}
\label{subsec:3}
Although we achieve accurate $\text{D}_d$ alignment, longer reasoning chains may introduce computational redundancy. As identified in~\cite{sui2025stopoverthinkingsurveyefficient}, such ``overthinking" reduces inference efficiency without improving output quality.

\paragraph{Observation}
Through empirical analysis in Figure~\ref{length}, we observe a strong correlation between degradation intensity and required reasoning length, as: 
\begin{equation}  \small
    \texttt{len}({\mathbf{Y}}) \propto \mathbb{E}\left[\sum_{i=1}^{I}s_d^{(i)}\right],
\end{equation}
where $\texttt{len}({\mathbf{Y}})$ denotes the length of the generated reasoning chain.
Higher degradation levels necessitate longer reasoning chains, while simpler degradations only require shorter responses.

\paragraph{Reward for Suitable Reasoning Length}
To optimize computational efficiency while maintaining robustness, we introduce a length-modulation reward (Figure~\ref{method}-B (right)):
\begin{equation} \small
r_{\text{len}}(\mathbf{Y}, \mathbf{Y}_{\text{GT}}) = 1 - \frac{| \texttt{len}(\mathbf{Y}) - \texttt{len}(\mathbf{Y}_{\text{GT}}) |}{\texttt{len}(\mathbf{Y}_{\text{GT}})},
\end{equation}
where $\texttt{len}(\mathbf{Y}_{\text{GT}})$ is the optimal length from ground truth. This reward equals $1$ when lengths match exactly $\texttt{len}(\mathbf{Y}) = \texttt{len}(\mathbf{Y}_{\text{GT}})$, and decreases linearly with relative length discrepancy.


\paragraph{Reinforcement Learning (RL)}
We integrate these two rewards into a unified optimization framework:
\begin{equation}  \small
\mathcal{R}(\mathbf{Y}, \mathbf{Y}_{\text{GT}}) = r_{\text{deg}}(\mathbf{Y}, \mathbf{Y}_{\text{GT}}) + r_{\text{len}}(\mathbf{Y}, \mathbf{Y}_{\text{GT}}),
\end{equation}
where $\mathcal{R}(\cdot)$ represents the comprehensive reward function.
This composite reward drives Group Relative Preference Optimization (GRPO)~\cite{shao2024deepseekmath}, and for each input pair $ \mathbf{X}_d \oplus \mathbf{P}$, we sample $G$ candidate responses $\{\mathbf{Y}^{(g)}\}_{g=1}^G$. The group-relative advantage is computed as:
\begin{equation}  \small
\hat{A}^{(g)} = \frac{\mathcal{R}^{(g)} - \mu_{\mathcal{R}}}{\sigma_{\mathcal{R}}},
\end{equation}
where $\mathcal{R}^{(g)} = \mathcal{R}(\mathbf{Y}^{(g)}, \mathbf{Y}_{\text{GT}})$, with:
\begin{equation}  \small
    \mu_{\mathcal{R}} = \frac{1}{G} \sum_{g=1}^{G} \mathcal{R}^{(g)}, \
\sigma_{\mathcal{R}} = \sqrt{ \frac{1}{G} \sum_{g=1}^{G} \left( \mathcal{R}^{(g)} - \mu_{\mathcal{R}} \right)^2 },
\end{equation}
Through GRPO optimization~\cite{shao2024deepseekmath}, we maximize the expected composite reward:
\begin{equation} \small
\theta^* = \arg\max_{\theta} \mathbb{E}_{(\mathbf{X}_d,\mathbf{P}) \sim \mathcal{P}_{T}} \left[ \mathcal{R}\left(\mathbf{Y}, \mathbf{Y}_{\text{GT}} \right) \right].
\end{equation}
This optimization strategy achieves dual objectives: (1) accurate alignment with degradation parameters through $r_{\text{deg}}$, and (2) suitable allocation of computational efficiency through $r_{\text{len}}$. The combined approach ensures robust visual understanding while maintaining efficiency across diverse real-world degradation scenarios.

\begin{figure*}
    \centering
    \includegraphics[width=1\linewidth]{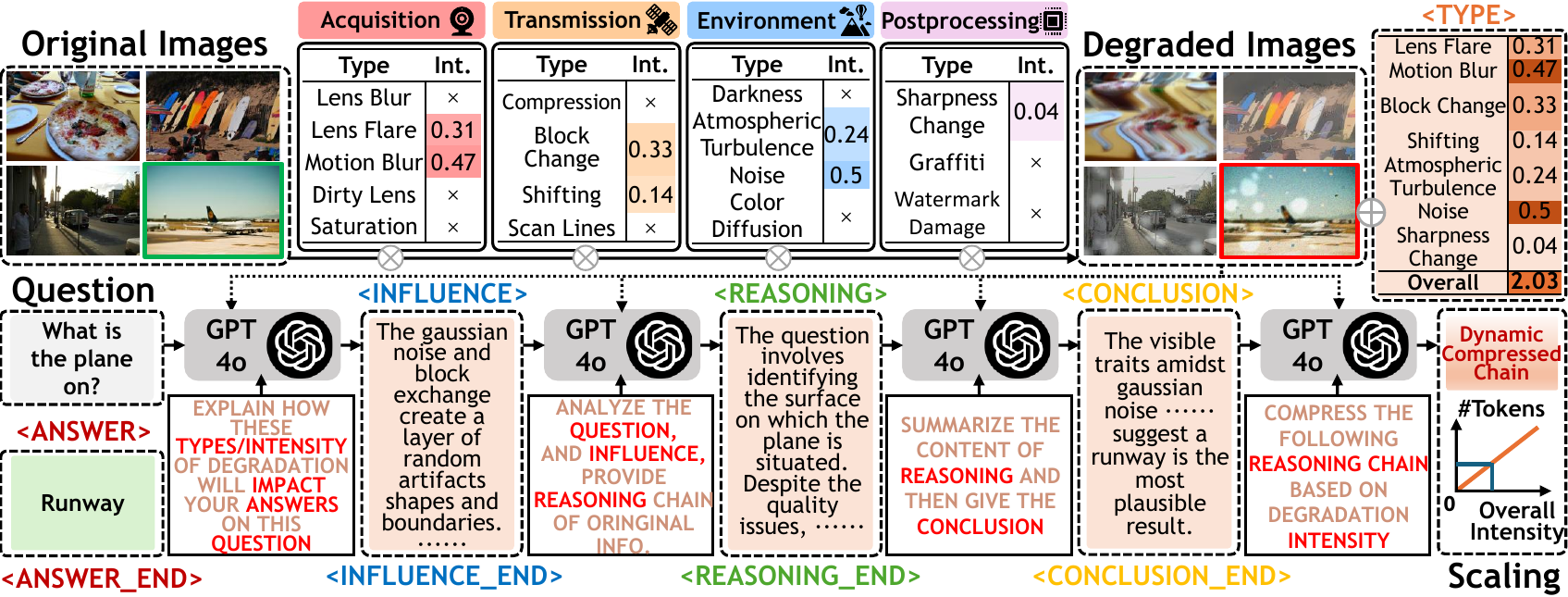}
    \caption{Data generation pipeline. The \textcolor{ForestGreen}{original images} undergo various real-world processing stages, where multiple degradations are randomly added to obtain \textcolor{red}{degraded images} and their corresponding degradation \texttt{<TYPE>}s. Based on these and the original question-answering pairs (QAs), the pipeline progressively generates \texttt{<INFLUENCE>}, \texttt{<REASONING>}, and \texttt{<CONCLUSION>}. Finally, the reasoning chain is scaling according to different \textbf{intensities} to achieve optimal efficiency.}
    \label{datagenpipiline}
\end{figure*}

\section{Data Construction} \label{dataset}
Existing visual understanding datasets (e.g., LLaVA~\cite{liu2023improvedllava}, R-Bench~\cite{li2024rbench}, A-OKVQA~\cite{schwenk2022aokvqabenchmarkvisualquestion}, Conceptual Captions~\cite{sharma2018conceptual}) lack explicit annotations for degradation parameters ($\text{D}_d$), their impacts ($\Delta_d$), and pristine semantic reasoning chains ($\text{T}_{\mathbf{X}}$).
This gap hinders training degradation-aware MLLMs.
To bridge this gap, we construct a specialized dataset featuring synthetically generated degradations and structured reasoning annotations. Our dataset is built upon a subset of A-OKVQA~\cite{schwenk2022aokvqabenchmarkvisualquestion}, comprising 10K samples for training and 1K for validation.

Our whole automated annotation pipeline, illustrated in Figure~\ref{datagenpipiline}. The procedure consists of the following five steps:
\paragraph{Step (1): Synthesizing Real-World Degradations}
We construct a comprehensive degradation model $\mathcal{D}(\cdot)$ that simulates degradations introduced across four real-world image processing stages:
\textbf{1. Acquisition} (Lens Blur, Lens Flare, Motion Blur, Dirty Lens, Saturation), \textbf{2. Transmission} (Compression, Block Change, Shifting, Scan Lines), \textbf{3. Environment} (Darkness, Atmospheric Turbulence, Noise, Color Diffusion), and \textbf{4. Postprocessing} (Sharpness Change, Graffiti, Watermark Damage).

For each pristine image $\mathbf{X}$, we generate a degraded version by:
\begin{equation} \small
\mathbf{X}_d = \mathcal{D}\left(\mathbf{X}\ ; \ \{\tau_d^{(i)}, s_d^{(i)}\}_{i=1}^{I}\right),
\end{equation}
where the degradation function $\mathcal{D}(\cdot)$ is parameterized by randomly sampled types $\tau_d^{(i)}$ and intensities $s_d^{(i)} \sim \mathcal{U}[0,1]$.

\paragraph{Step (2): Generating Degradation Influence}
We employ GPT-4o~\cite{hurst2024gpt4o} with a fixed prompt template $\Psi_{\text{INFLUENCE}}$ to produce a textual description $\Delta_d$ of the degradation's semantic impact:
\begin{equation}  \small
\Delta_d = \mathcal{G}_{\text{GPT-4o}}(\mathbf{X}, \mathbf{X}_d, \text{D}_d, \mathbf{Y}_{\text{GT}} \ ;\ \Psi_{\text{INFLUENCE}}).
\end{equation}
This narrative establishes a causal link between the visual degradation and its effect on content interpretation, providing the necessary supervision for training the perception module $\mathcal{M}_{\text{p}}(\cdot)$.

\paragraph{Step (3): Generating Pristine Semantic Reasoning}
Using a distinct prompt template $\Psi_{\text{REASONING}}$, we instruct GPT-4o to infer the original semantic reasoning chain $\text{T}_{\mathbf{X}}$ by compensating for the degradation influence:
\begin{equation}  \small
\text{T}_{\mathbf{X}} = \mathcal{G}_{\text{GPT-4o}}(\mathbf{X}_d, \text{D}_d, \Delta_d, \mathbf{Y}_{\text{GT}}\ ;\ \Psi_{\text{REASONING}}),
\end{equation}
This step recovers the underlying reasoning process as if performed on the undistorted image, which is crucial for training the reconstruction module $\mathcal{M}{\text{r}}(\cdot)$.

\paragraph{Step (4): Generating Reasoning Conclusion} The final reasoning conclusion $\mathbf{Y}_d$ is generated by conditioning on the pristine semantic reasoning and the ground-truth answer, using a prompt template $\Psi_{\text{CONCLUSION}}$:
\begin{equation}  \small
\mathbf{Y}_d = \mathcal{G}_{\text{GPT-4o}}(\text{T}_{\mathbf{X}}, \mathbf{Y}_{\text{GT}}\ ;\ \Psi_{\text{CONCLUSION}}).
\end{equation}

\paragraph{Step (5): Scaling Reasoning Chain Length}
To enable adaptive computational allocation, we dynamically adjust the length of the complete reasoning chain $\text{C}$ based on the total degradation intensity:
\begin{equation}  \small
\hat{\text{C}} = \mathcal{G}_{\text{GPT-4o}} \left( \text{C}\ ;\ \Psi_{\text{Len}} (\sum_{i=1}^{I} s_d^{(i)}) \right),
\end{equation}
where $\hat{\text{C}}$ denotes the scaled reasoning chain, and $\Psi_{\text{Len}}(\cdot)$ is a set of intensity-calibrated prompt templates. This procedure ensures reasoning efficiency and is instrumental for optimizing the length reward $r_{\text{len}}$.

\paragraph{ Quality and Robustness}
The resulting dataset, structured according to the reasoning process defined in Eq.~\eqref{chain}, supports both the SFT and the subsequent GRPO optimization of our robust model $\mathcal{M}_{\text{MLLM}}^{\text{(Robust)}}(\cdot)$.
Besides, the inverse relation between image quality and degradation intensity validates that the distribution of corruptions in our dataset mirrors real-world conditions.
The lexical diversity of the reasoning corpus, demonstrates its inherent capacity to model complex logical relationships. This establishes a foundation for achieving robust performance. \textit{More details in the supplementary material}.

\begin{table*}[t]
    \centering
    \resizebox{\linewidth}{!}{
    \begin{tabular}{c|l|ccc|ccc|ccc|c}
    \toprule
    \multirow{2.5}{*}{\textbf{Category}}      & \multirow{2.5}{*}{\textbf{Method}}          & \multicolumn{3}{c|}{\textbf{MCQ}}       & \multicolumn{3}{c|}{\textbf{VQA}}       & \multicolumn{3}{c|}{\textbf{CAP}}       & \multirow{2.5}{*}{\textbf{Overall}} \\
    \cmidrule{3-5} \cmidrule{6-8} \cmidrule{9-11}
                 &                 & low       & mid       & high      & low       & mid       & high      & low       & mid       & high      &                          \\
    \midrule
    \multirow{3}{*}{\shortstack{General \\ MLLM}}
                & Qwen2.5-VL-3B~\cite{Qwen2.5-VL}      & \textcolor{blue}{0.6411}& 0.6022      & \textcolor{blue}{0.5732}      & 0.4872      & \textcolor{blue}{0.4854}      & 0.4904      & 0.3778      & 0.3704      & 0.3330      & 0.4845      \\
                 & Gemma3-4B~\cite{team2025gemma}      & 0.5823      & 0.5776      & 0.5060      & 0.4865      & 0.4630      & 0.4419      & 0.4048      & 0.3746      & 0.3480      & 0.4649      \\
                 & InternVL-4B~\cite{chen2024internvl}     & 0.6235      & 0.6024      & 0.5914      & \textcolor{red}{0.4982}      & 0.4539      & 0.5108      & 0.3667      & 0.3041      & 0.2851      & 0.4706      \\
    \midrule
    \multirow{3}{*}{\shortstack{Robust\\ MLLM}}
                    & TeCoA~\cite{wang2024tecoa} & 0.4647 & 0.4223 & 0.4024 & 0.4687 & 0.3994 &0.4461 &0.2111 & 0.2195 & 0.1937 & 0.3586  \\
                    & Robust CLIP~\cite{schlarmann2024robustclip} & 0.4705 & 0.4658 & 0.4024 & 0.4503 &0.4339& 0.4743& 0.2290& 0.2219&0.1983 & 0.3718 \\
                    & Robust LLaVA~\cite{malik2025robust} & 0.3352 & 0.2608 & 0.3048 & 0.2607 & 0.2212 & 0.2443 & 0.0068 & 0.0065 & 0.0067 & 0.1830 \\
    \midrule 
    \multirow{2}{*}{{Ours}}
                & SFT           & 0.6176 & \textcolor{blue}{0.6087}& 0.5610 & 0.4804 & 0.4836 & \textcolor{red}{0.5012} & \textcolor{red}{0.4080} & \textcolor{red}{0.3858} & \textcolor{red}{0.3518} & \textcolor{blue}{0.4886} \\
                 & SFT and RL   & \textcolor{red}{{0.6529}} & \textcolor{red}{{0.6391}} & \textcolor{red}{{0.6097}} & \textcolor{blue}{0.4914} & \textcolor{red}{0.4909} & \textcolor{blue}{0.4980} & \textcolor{blue}{0.4068} & \textcolor{blue}{0.3781} & \textcolor{blue}{0.3484} & \textcolor{red}{0.5017} \\
    \bottomrule                                 
    \end{tabular}}
    \caption{Quantitative performance on R-Bench~\cite{li2024rbench} on MCQ/VQA/CAP tasks with three degradation strength levels (from low to high). 
    The best/second best results are shown in \textcolor{red}{Red}/\textcolor{blue}{{Blue}} respectively.}
    \label{rbench}
\end{table*}

\begin{table*}[t]
    \centering
    \resizebox{\linewidth}{!}{
    \begin{tabular}{c|l|cccc|cccc|cccc}
    \toprule
    \multirow{3.5}{*}{\textbf{Category}}      & \multirow{3.5}{*}{\textbf{Method}}          & \multicolumn{4}{c|}{\textbf{MMMB}~\cite{sun2025parrotmultilingualvisualinstruction}}       & \multicolumn{4}{c|}{\textbf{MMStar}~\cite{chen2024we}}       & \multicolumn{4}{c}{\textbf{RealWorldQA}~\cite{xai2024grok15v}}        \\
    \cmidrule{3-6} \cmidrule{7-10} \cmidrule{11-14}
                 &                 & \multirow{2}{*}{clean}       & \multicolumn{3}{c|}{Intensity}       & \multirow{2}{*}{clean}       & \multicolumn{3}{c|}{Intensity}  & \multirow{2}{*}{clean}       & \multicolumn{3}{c}{Intensity}  \\
                &                 &     & $25\%$ & $50\%$ & $100\%$       &        & $25\%$ & $50\%$ & $100\%$  &        & $25\%$ & $50\%$ & $100\%$   \\
                 
    \midrule
    \multirow{3}{*}{\shortstack{General \\ MLLM}}
                & Qwen2.5-VL-3B~\cite{Qwen2.5-VL}    & 80.60 & 79.19 & \textcolor{blue}{78.68} & 74.50 & 54.73& 52.90 & \textcolor{blue}{51.86} & \textcolor{blue}{48.66} & 65.22 & 64.96 & 63.39 & 60.65 \\
                 & Gemma3-4B~\cite{team2025gemma}    & 71.01 & 70.30& 70.20 & 69.14 & 43.93 & 43.20 & 42.60 & 41.33  & 55.42 & 54.77 &  53.72 & 52.81 \\
                 & InternVL-4B~\cite{chen2024internvl}    & 77.97 & 77.47 & 76.66 & 74.59 & 51.53 & 50.26 & 49.60 & 46.93 & 57.38 & 58.16 & 57.64 &  54.90 \\
    \midrule
    \multirow{2}{*}{\shortstack{Robust\\ MLLM}}
                    & TeCoA~\cite{wang2024tecoa} & 57.17 & 65.71 & 56.11 & 51.76  & 30.46 & 30.60 & 30.73 & 28.06 & 40.00 & 39.73 & 39.47 & 38.69 \\
                    & Robust CLIP~\cite{schlarmann2024robustclip} & 58.83 & 58.28 & 57.97 & 53.33 &33.00& 32.26 & 31.80 & 29.46 &43.26& 42.48& 42.61 & 41.43\\
    \midrule
    \multirow{2}{*}{{Ours}}
                & {SFT}          & \textcolor{blue}{80.85} & \textcolor{blue}{79.45} & \textcolor{blue}{78.68} & \textcolor{blue}{74.94} & \textcolor{blue}{55.20} & \textcolor{blue}{53.00} & \textcolor{blue}{51.86} & \textcolor{red}{49.53} & \textcolor{red}{68.23} & \textcolor{red}{67.58}  & \textcolor{red}{67.32} & \textcolor{red}{63.92} \\
                 & {SFT and RL}        & \textcolor{red}{81.41} & \textcolor{red}{79.49} & \textcolor{red}{79.04} & \textcolor{red}{75.35} & \textcolor{red}{56.86} & \textcolor{red}{54.40} & \textcolor{red}{53.60} & \textcolor{red}{49.53} & \textcolor{blue}{67.71}  & \textcolor{blue}{66.40} & \textcolor{blue}{67.05} & \textcolor{blue}{63.26} \\
    \bottomrule                                 
    \end{tabular}}
    \caption{Quantitative performance for anti-degradation on three visual understanding benchmarks (MMMB~\cite{sun2025parrotmultilingualvisualinstruction}, MMStar~\cite{chen2024we}, and RealWorldQA~\cite{xai2024grok15v}) with three degradation intensity levels (from 25\% to 100\%). 
    The best/second best results are showed in \textcolor{red}{{Red}}/\textcolor{blue}{{Blue}} respectively.}
    \label{anti-degradation}
\end{table*}

\section{Experiments}

\paragraph{Training Configuration}
Our model is built upon Qwen2.5-VL-3B~\cite{Qwen2.5-VL}, which employs a redesigned Vision Transformer (ViT) as its vision encoder. We adopt a dual-stage optimization strategy:
\begin{itemize}
    \item \textbf{Supervised Fine-Tuning (SFT)}: $25\%$ training data used to establish basic instruction-following ability.
    \item \textbf{Reinforcement Learning (RL)}: $75\%$ data for align accurate degradation parameters and suitable chain length.
\end{itemize}
Notably, we freeze both the vision encoder and visual projection layers while performing \textit{full-parameter fine-tuning} on the language model.
This design preserves visual feature stability while empowering the MLLM to develop robust degradation-aware reasoning mechanisms.

\paragraph{Baselines}
We compare against two categories SOTA baselines: (i) \textbf{General MLLMs}, including Qwen2.5-VL-3B~\cite{Qwen2.5-VL}, Gemma3-4B~\cite{team2025gemma}, and InternVL-4B~\cite{chen2024internvl}; (ii) \textbf{Robust MLLMs}, comprising TeCoA~\cite{wang2024tecoa}, Robust CLIP~\cite{schlarmann2024robustclip}, and Robust LLaVA~\cite{malik2025robust}.

\paragraph{Benchmarks}
We conduct rigorous evaluation across two dimensions: (i) \textbf{Real-World Robustness}: Directly assessing robust visual understanding ability on R-Bench~\cite{li2024rbench};
(ii) \textbf{Adversarial Robustness}: Evaluation under synthetic degradation attacks by applying multi-type, multi-level real-world degradations to visual content in MMMB~\cite{sun2025parrotmultilingualvisualinstruction}, MMStar~\cite{chen2024we}, and RealWorldQA~\cite{xai2024grok15v}.
This dual-strategy comprehensively measures both intrinsic degradation comprehension and performance preservation under visual corruption.

\begin{figure}
    \centering
    \includegraphics[width=1\linewidth]{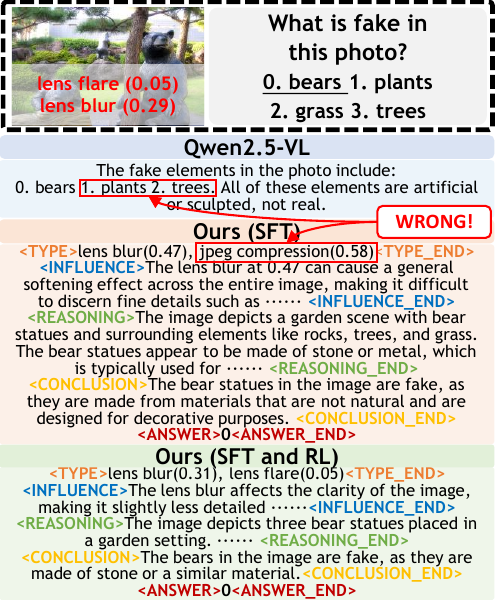}
    \caption{Qualitative evaluation for anti-degradation. Ours (SFT and RL) can provide robust and efficient result.}
    \label{visual}
\end{figure}

\begin{table*}[t]
    \centering
    \resizebox{\linewidth}{!}{
    \begin{tabular}{l|ccc|ccc|ccc|c}
    \toprule
     \multirow{2.5}{*}{\textbf{Method}}   & \multicolumn{3}{c|}{\textbf{MCQ}}       & \multicolumn{3}{c|}{\textbf{VQA}}       & \multicolumn{3}{c|}{\textbf{CAP}}       & \multirow{2.5}{*}{\textbf{Overall}} \\
    \cmidrule{2-4} \cmidrule{5-7} \cmidrule{8-10}
                        & low       & mid       & high      & low       & mid       & high      & low       & mid       & high      &                          \\
    \midrule
                 Qwen2.5-VL-3B~\cite{Qwen2.5-VL}    & 0.6411      & 0.6022      & 0.5732      & 0.4872      & 0.4854      & 0.4904      & \textcolor{blue}{0.3778}      & \textcolor{blue}{0.3704}      & \textcolor{blue}{0.3330}      & 0.4845      \\
    \midrule
                 {Ours} (w/o Reasoning) & \textcolor{blue}{0.6588} & 0.5901 & 0.4756 & 0.4905 & \textcolor{blue}{0.4900} & 0.4862 & 0.2901 & 0.2673 & 0.2758 & 0.4471 \\
                 {Ours} (w/o $r_{\text{deg}}$)  & \textcolor{red}{0.6647} & \textcolor{red}{0.6398} & 0.5505 & \textcolor{blue}{0.4912} & 0.4894 & \textcolor{red}{0.5056} & 0.3684 & 0.3578 & 0.3248 & 0.4880 \\
                 {Ours} (w/o $r_{\text{len}}$)  &\textcolor{red}{0.6647} & 0.6354 & 0.5975 & 0.4904 & 0.4887 & 0.4877 & 0.3656 & 0.3678 & 0.3189 & \textcolor{blue}{0.4907} \\
    \midrule
                 {Ours}           & 0.6529 & \textcolor{blue}{0.6391} & \textcolor{red}{0.6097} & \textcolor{red}{0.4914} & \textcolor{red}{0.4909} & \textcolor{blue}{0.4980} & \textcolor{red}{0.4068} & \textcolor{red}{0.3781} & \textcolor{red}{0.3484} & \textcolor{red}{0.5017} \\
    \bottomrule                                 
    \end{tabular}}
    \caption{Ablation study on R-Bench~\cite{li2024rbench} on MCQ/VQA/CAP tasks with three degradation strength levels (from low to high). 
    The best/second best results are showed in \textcolor{red}{{Red}}/\textcolor{blue}{{Blue}} respectively.}
    \label{ablation}
\end{table*}

\subsection{Performance on R-Bench}
R-Bench~\cite{li2024rbench} is a benchmark designed to directly evaluate image understanding capabilities under real-world degradation conditions. It incorporates three distinct task types (Multiple Choice Questions (MCQ), Visual Question Answering (VQA), and Image Captioning (CAP)) with three degradation intensity levels (low, mid, and high) to systematically assess the robustness of visual comprehension.

As shown in Table~\ref{rbench}, \name\ (Ours) demonstrates significant improvements in image understanding capabilities following both Supervised Fine-Tuning (SFT) and subsequent Reinforcement Learning optimization (SFT and RL). Experimental results indicate that our model surpasses existing general and robust MLLMs baselines in overall performance on this benchmark.

\subsection{Anti-Degradation Performance}
To rigorously evaluate our model's robustness against image degradation, we conduct comprehensive experiments on three established visual understanding benchmarks (MMMB~\cite{sun2025parrotmultilingualvisualinstruction}, MMStar~\cite{chen2024we}, and RealWorldQA~\cite{xai2024grok15v}). We introduce random degradations at varying intensity levels (25\%, 50\%, and 100\%) to the original images, creating challenging test conditions that assess the model's anti-degradation capability.

\paragraph{Quantitative Results}
As demonstrated in Table~\ref{anti-degradation}, our model achieves SOTA performance across all degradation levels compared to existing baselines.
This evidence confirms our model's exceptional robustness to diverse image degradations under adversarial conditions.

\paragraph{Qualitative Result} Figure~\ref{visual} presents qualitative comparisons of our outputs. Compared to the original baseline, \name\ significantly reduces hallucinations and errors in visual understanding through reasoning.
Furthermore, after preference optimization, \name\ achieves a optimal balance between inference efficiency and accurate degradation parameters perception.

\subsection{Ablation Study} 
\paragraph{Reasoning \textit{vs.} Adaptation}
To validate the effectiveness of explicit reasoning versus implicit adaptation, we conduct an ablation study by removing degradation reasoning chains from our training data, relying solely on fine-tuning for adaptation (Table~\ref{ablation}, w/o Reasoning). The experimental results reveal two critical findings:
(i) Adaptation provides only marginal performance gains in specific intensity ranges compared to the base model, and fails catastrophically in high-intensity degradation scenarios;
(ii) Explicit reasoning demonstrates significantly improved robustness over both the adaptation-only model and the original baseline.
These results conclusively demonstrate that explicit reasoning capability is essential for robust visual understanding, enabling systematic analysis and compensation for visual degradations rather than mere adaptation.

\begin{figure}
    \centering
    \includegraphics[width=1\linewidth]{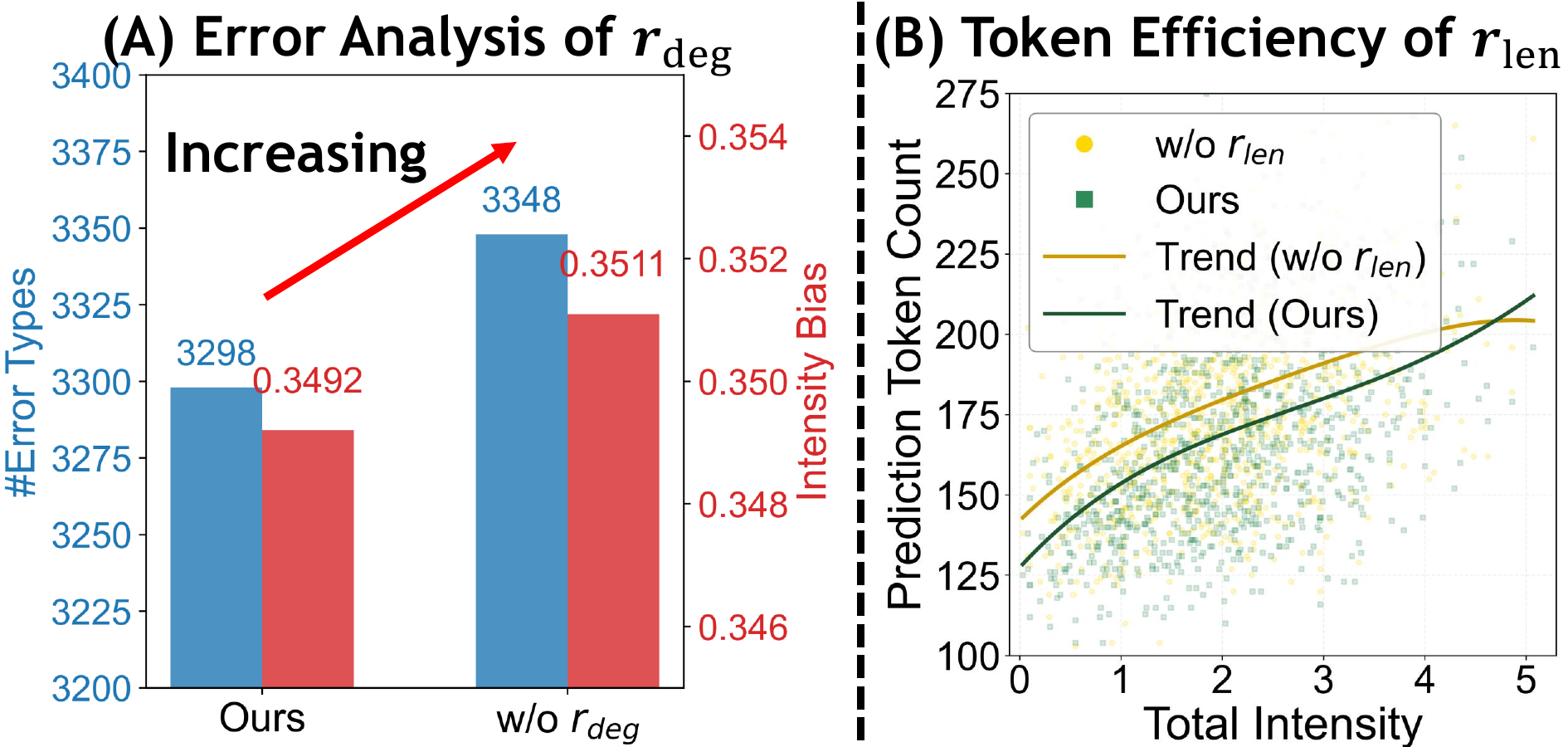}
    \caption{Statistics analysis for (A) $r_{\text{deg}}$ and (B) $r_{\text{len}}$.}
    \label{ablationstat}
\end{figure}

\paragraph{Effectiveness of $r_{\text{deg}}$}
To validate the critical role of the degradation reward $r_{\text{deg}}$, we conduct an ablation study comparing model performance with and without this component. 
As shown in Table~\ref{ablation}, incorporating $r_{\text{deg}}$ substantially improves visual understanding performance on R-Bench compared to the ablated variant. This improvement stems from $r_{\text{deg}}$'s ability to enhance precise alignment with degradation parameters. 
Furthermore, statistical analysis on our out-of-domain testset (Section~\ref{dataset}) in Figure~\ref{ablationstat}-A reveals that $r_{\text{deg}}$ significantly reduces two key error types: (i) degradation-type misclassification and (ii) degradation-intensity estimation bias. These results demonstrate that $r_{\text{deg}}$ increases model precision in identifying degradation parameters, directly contributing to superior robustness.

\paragraph{Efficiency of $r_{\text{len}}$}
To evaluate the effectiveness of the length-modulation reward $r_{\text{len}}$, we conduct an ablation study by removing this component.
As shown in Figure~\ref{ablationstat}-B, incorporating $r_{\text{len}}$ reduces the average reasoning chain length while maintaining performance, demonstrating its ability to improve computational efficiency.
Notably, the model adaptively adjusts reasoning depth based on degradation intensity: longer chains are allocated for severe degradation, while simpler cases require less inference. 
This task-adaptive allocation not only optimizes resource usage but also enhances overall performance, as evidenced by the quantitative improvements in Table~\ref{ablation} (w/o $r_{\text{len}}$).

\section{Conclusion}
We propose \name, a novel paradigm that incorporates explicit degradation reasoning chains to enhance multimodal understanding robustness. 
We believe this work opens new avenues for building more robust, interpretable, and efficient multimodal systems capable of operating reliably in visually challenging environments.

\section*{Acknowledgments}
The work described in this paper was supported by a grant from the Research Grants Council of the Hong Kong Special Administrative Region, China (Project Reference Number: AoE/E-601/24-N).

Besides, this work was supported in part by the National Natural Science
Foundation of China (No. 62472359, 62372379), in part by the Xi’an’s Key
Industrial Chain Core Technology Breakthrough Project: AI Core Technology
Breakthrough under Grand 24ZDCYJSGG0003.
Also, this work was supported by the Key Project of the National Natural Science Foundation of China (No. 62536007), the Zhejiang Province Science Foundation (No. LD24F020002) and the Zhejiang Province's 2025 "Leading Goose + X" Science and Technology Plan (No. 2025C02034). 

\bibliography{aaai2026}

\end{document}